# Introducing Answered with Evidence - a framework for evaluating whether LLM responses to biomedical questions are founded in evidence.

Julian D Baldwin[1], Christina Dinh[1], Arjun Mukerji[1], Neil Sanghavi[1], Saurabh Gombar[1,2]

## Author Affiliations

[1]Atropos Health; New York, NY, USA
[2]Stanford School of Medicine - Department of Pathology, Stanford University, Stanford, California, USA

## Abstract

The growing use of large language models (LLMs) for biomedical question answering raises concerns about the accuracy and evidentiary support of their responses. To address this, we present *Answered with Evidence*, a framework for evaluating whether LLM-generated answers are grounded in scientific literature. We analyzed thousands of physician-submitted questions using a comparative pipeline that included: (1) Alexandria™, fka the Atropos Evidence™ Library, a retrieval-augmented generation (RAG) system based on novel observational studies, and (2) two PubMed-based retrieval-augmented systems (System and Perplexity). We found that PubMed-based systems provided evidence-supported answers for approximately 44% of questions, while the novel evidence source did so for about 50%. Combined, these sources enabled reliable answers to over 70% of biomedical queries. As LLMs become increasingly capable of summarizing scientific content, maximizing their value will require systems that can accurately retrieve both published and custom-generated evidence—or generate such evidence in real time.

## Background

Evidence-based medicine—making clinical decisions based on the best available research—has been the standard of care for over three decades[1]. However, in some specialties, fewer than 20% of daily medical decisions are supported by high-quality evidence[2]. This gap arises primarily from two challenges: First, clinical trials often lack generalizability to real-world patients, especially those with complex comorbidities who are frequently excluded from study populations[3]. Second, even when relevant studies exist, their findings may conflict due to heterogeneity in patient populations, variable study quality, and inconsistent endpoints and difficulty in the physician to access the evidence[4]. These limitations hinder the synthesis of clear, actionable recommendations for individual patients, leading clinicians to rely heavily on summarized evidence from trusted sources[5].

Large language models (LLMs) are increasingly being explored for biomedical applications, including literature summarization and natural language question answering[6–8]. LLMs have demonstrated impressive capabilities in extracting and synthesizing biomedical information, but they are also prone to hallucinating sources, fabricating citations, or generating responses that deviate from established clinical guidelines[9–11]. These shortcomings present risks in domains where accuracy is critical. Despite these concerns, LLM integration into biomedical research and clinical workflows has accelerated markedly in recent years.

One emerging use case is the use of LLMs to generate evidence summaries to support clinical decision-making. Several commercial services—such as Open Evidence, System, Consensus, SciSummary, Perplexity, and others—now offer LLM-powered interfaces to biomedical literature [12,13]. While adoption of these tools is increasing, questions remain about how to evaluate their reliability and trustworthiness. Frameworks like the recently proposed MedHELM [14,15] aim to standardize the evaluation of LLM outputs across tasks, but they do not yet assess whether responses are grounded in verifiable biomedical evidence.

In this study, we introduce *Answered with Evidence*, a structured methodology for evaluating the correctness, evidentiary grounding, and citation fidelity of LLM-generated answers to biomedical questions. Our approach emphasizes verifiability against trusted biomedical sources, enabling the development and monitoring of systems suitable for clinical decision support and real-world evidence generation. To assess the utility of this framework, we analyzed nearly three-thousand physician-submitted questions to Atropos Health's Green Button Service [8,16,17]— a human-in-the-loop, real-world evidence platform used by clinicians, hospital leaders, and researchers. Answers were compared across two PubMed-based retrieval-augmented systems (System and Perplexity) and Alexandria, the Atropos Evidence Library, a curated collection of real-world evidence derived from prior medical inquiries.

## Materials and Methods

Figure 1 outlines the overall analysis workflow. We began by selecting a sample of 2,972 questions submitted to Atropos Health between 2022 and 2025. These questions originated from physicians seeking evidence to inform clinical decisions and from researchers evaluating treatment efficacy or event rates in defined populations. The sample spans all major clinical subspecialties and includes a diverse range of query types—from causal inference questions related to adverse events and treatment comparisons to more straightforward epidemiological questions about prevalence and incidence.

Each question was processed on an LLM with access to different underlying evidence sources or with different methodologies. Once the response was obtained it was evaluated for being based in evidence and then based on the evaluation a "badge" was assigned to ease in user interpretation.

## Evidence Sources

We evaluated biomedical question-answering systems using three distinct evidence sources, each optimized for different types of clinical inquiries. The first two sources—System Inc[13]. and Perplexity[12]—are retrieval-augmented generation[18] (RAG) platforms grounded in the PubMed corpus, comprising peer-reviewed biomedical literature that reflects the current state of scientific knowledge. Perplexity, while designed as a general-purpose LLM-based question-answering tool, was configured for this analysis to retrieve only from academic publications. SystemAI combines large language models with graph-based retrieval technologies, trained specifically on scientific literature using state-of-the-art models.

The third evidence source is Alexandria, the Atropos Evidence Library, a proprietary collection of structured real-world evidence (RWE). This library includes studies generated from electronic health records, open and closed claims data, and specialized clinical datasets in response to specific physician and researcher inquiries. Each study includes human-readable definitions of the population, cohorts, exposures, and outcomes. Collectively, these three sources provide a comprehensive basis for evaluating the correctness, evidence alignment, and clinical relevance of LLM-generated answers.

Each question was processed using two evidence sources: Alexandria, the Atropos Evidence Library and the System platform. A subset of 1,739 questions was randomly selected for evaluation using the Perplexity platform. This subset size was chosen to balance computational cost and runtime constraints while maintaining sufficient diversity to represent the broader range of biomedical queries.

## Evidence Evaluation

We apply a structured evaluation rubric composed of three binary (True/False) criteria. Each criterion isolates a specific dimension of answer quality with respect to its contextual support.

| Criterion | Definition | True | False |
| --- | --- | --- | --- |
| Context directly answers the question | Evaluates whether the provided context includes specific, relevant information that fully or substantially answers the question | The context contains facts, statistics, outcomes that clearly respond to the core question | The context lacks the information needed to directly address the question |
| Context is related to | Evaluates whether | The context | The context is |

| the question | the provided context is semantically or topically related to the question, even if it doesn't contain a direct answer | discusses the same disease, drug, intervention, outcome as the question | unrelated or discusses an entirely different clinical scenario |
| Answer is grounded in the provided context | Evaluates whether the AI-generated answer relies solely on the provided context without introducing unsupported claims, hallucinated facts, or external knowledge | The answer faithfully summarizes, paraphrases, or infers only from the information found in the context | The answer adds information not present in the context or contradicts it |

Once the appropriate context is retrieved and an AI-generated answer returned, we use a structured prompt to judge the answers' relevance and grounding. This evaluation assesses how well the AI-generated answer responds to the original question, given only the retrieved context (Full prompt details available in Supplemental Figure S1)

To simplify interpretation, each evaluated question-answer pair is assigned a color-coded Answered with Evidence badge based on the boolean fields from the structured prompt output. An example of the evaluation of a question-response context and badge assignment is shown in Figure 2.

| **Badge Color** | **Context directly answers the question** | **Context is related to the question** | **Answer is grounded in the provided context** |
|---|---|---|---|
| Green | True | True | True |
| Yellow | False | True | True |
| Red | Any other combination | | |

This logic provides a quick way to assess the quality and reliability of evidence grounding in biomedical question-answer responses. The badge system is designed to classify the factual alignment of AI-generated answers and signal the evidence landscape surrounding a biomedical question.The green badge indicates a high-confidence response. These answers represent areas of well-established evidence where the AI-generated answer is effectively synthesizing known information. The yellow badge indicates the context is relevant to the question, and the answer is grounded in it, but the context does not directly answer the question. It reflects situations where indirect evidence or background knowledge exists, but no study has yet addressed the specific question. This marks the evidence gap and the opportunity to generate a novel study. The red badge indicates the context is not related, the answer introduces hallucinated content, or both. It represents a low-confidence response.

## Results

### Evidence Criterion Assessment and Answered with Evidence Badge assignment across evidence sources.

All questions were submitted to the three evaluated evidence sources, and each response was assessed based on the criteria described in the Evidence Evaluation section. A summary of the badge assignment for the three evidence sournces can be found on Table 1.For the PubMed-based sources, the System platform produced high-confidence, evidence-grounded responses (Green badge) for 44.6% of queries. Perplexity, when restricted to academic literature, achieved Green-badged responses for 21.3% of cases. The remaining responses from these sources were assigned Yellow or Red badges, indicating that either the answer lacked sufficient evidence grounding (Yellow) or was irrelevant to the original question (Red). These lower-confidence responses comprised 55.4% for System and 78.7% for Perplexity.

In comparison, Alexandria, the Atropos Evidence Library, which generates responses using structured real-world evidence from EHR, claims, and other clinical datasets, produced Green-badged responses for 50.1% of the submitted questions. The remaining 49.9% were rated Yellow or Red, typically due to limitations in available real-world data or ambiguity in the question framing.

These findings suggest that novel real-world evidence sources may provide comparable or improved grounding relative to PubMed-based retrieval systems, particularly for complex or underrepresented clinical scenarios.

### Intersource agreement for Evidence Badge assignment

To assess the consistency of evidence grounding, we compared badge assignments across the three evidence sources for the subset of questions answered by multiple systems. Our hypothesis was that the two PubMed-based sources (System and Perplexity) would show higher concordance in badge assignment compared to the custom real-world evidence source (Alexandria), given their shared reliance on published literature. We also anticipated only partial agreement across sources due to differences in retrieval mechanisms and response generation strategies. Results are summarized in Table 2a–d.

Between System and Perplexity, Green badge agreement—signifying that both systems independently generated evidence-grounded responses—occurred in 11.1% of cases.

Concordance on Yellow or Red badges, indicating responses that were partially supported or not relevant, was 45.3%. Overall agreement across all badge categories was 56.3%.

Comparing Alexandria to Perplexity, Green badge agreement was 11.3%, while Yellow/Red agreement was 45.1%, resulting in an overall agreement rate of 56.4%.

In the comparison between Alexandria and System, Green badge agreement was higher at 21.9%, while Yellow/Red agreement was 32.5%. Overall agreement across this pair was 54.4%.

When considering agreement across **all three sources**, Green badge concordance was observed in just 6.3% of cases, while agreement on Yellow or Red responses occurred in 27.3% of cases. However, at least one evidence source provided a Green-badged answer in 72.7% of the cases, highlighting the complementary nature of these platforms.

To assess each source's unique contribution, we also measured *novelty*—defined as the proportion of questions where only one source produced a Green-badged response, while the others returned Yellow or Red. Novelty rates were 17.9% for Alexandria, 17.8% for System, and 5.2% for Perplexity. These results suggest that real-world evidence sources and PubMed-based systems each provide distinct and valuable contributions to biomedical question answering.

## Discussion

One of the enduring challenges in clinical care and health policy is the ability to access evidence that directly addresses the question at hand[19,20]. Despite the widespread adoption of evidence-based medicine, physicians and decision-makers often face a gap between available evidence and real-world information needs. In part, this challenge stems from the sheer volume and fragmentation of biomedical literature, which far exceeds the capacity of any individual to comprehensively access, evaluate, and synthesize[2,4,5].

The emergence of large language models (LLMs) with capabilities in literature retrieval and summarization offers a promising path forward[6]. However, even with accurate identification and synthesis of relevant publications, there remains a persistent evidence gap—many questions simply fall outside the scope of existing published studies. Bridging this gap requires not only tools to extract and summarize existing literature, but also systems that can surface or generate novel real-world evidence in response to specific clinical or research queries[16,17].

In this study, we introduced *Answered with Evidence*, a framework for evaluating whether LLM-generated responses are (1) relevant, (2) directly responsive to the question, and (3) grounded in evidence. Each response is scored across these criteria and summarized using an intuitive traffic light–style badge system—Green (high confidence and evidence-based), Yellow (partial or uncertain grounding), and Red (unsupported or irrelevant). This system provides users with rapid, interpretable feedback on the reliability of an answer.

Across the three evaluated evidence sources, Alexandria, the Atropos Evidence Library achieved the highest rate of Green-badged responses (50.1%), compared to 44.6% and 21.3% for System and Perplexity, respectively. Alexandria and System also demonstrated the highest *novelty* rates—defined as the proportion of Green-badged answers not replicated by the other sources—at 17.9% and 17.8%, respectively, versus 5.2% for Perplexity. These findings underscore the value of combining both published and real-world evidence in biomedical question answering.

Notably, despite both being grounded in PubMed literature, System and Perplexity exhibited markedly different performance profiles. The discrepancy in Green badge rates and novelty between these two systems suggests that training data alone does not fully explain differences in response quality. Architecture, retrieval mechanisms, and response-generation strategies also likely play a significant role.

Assuming continued improvements in LLM retrieval fidelity and hallucination reduction, we expect PubMed-based systems to eventually converge toward similar badge distributions—potentially reaching a ceiling of around 55% Green-badge responses. In this context, the ability to access or generate *novel* evidence becomes essential. In our analysis, supplementing Perplexity or System with Alexandria increased the proportion of high-confidence responses by an additional 25.7%–33.5%, demonstrating the unique value of real-world evidence in addressing unanswered clinical questions. If we just isolate System as the literature based approach with the most green badges, and Alexandria we find that no single source has more than 50% green-badges but taken together at least one source produced a green badged result 70.2% of the time (Figure 3)

This study has several limitations. First, the evaluation relies on a single LLM-as-a-judge to assess response quality, which, while efficient and interpretable, may introduce model-specific bias or inconsistency. Future work should explore an LLM-as-a-jury approach to improve robustness through ensemble evaluation. Second, several emerging LLM-based literature summarization platforms were not included in this initial analysis and warrant future benchmarking. Third, the evaluation prompt was designed to be general-purpose and not tailored to the response styles of specific LLMs. Given variations in hedging language and response structure, source-specific tuning may enhance the accuracy of badge classification.

## Conclusion

Accessing pertinent and reliable evidence remains a significant challenge for healthcare providers, health system leaders, and biomedical researchers. While large language models (LLMs) offer a powerful means to navigate the vast and expanding body of published literature, our findings demonstrate that even with advanced retrieval and summarization capabilities, approximately 55% of biomedical questions remain inadequately addressed using published sources alone. Similarly, a real-world evidence (RWE) library specifically designed to address gaps in the literature still left nearly half of questions unanswered. However, when combined, these complementary evidence sources enabled high-confidence answers for nearly 75% of clinician- and researcher-submitted inquiries.

These findings highlight a key insight: advancing biomedical question answering will require more than improved retrieval of existing literature—it will also require the generation and integration of novel evidence. To realize the full potential of LLMs in clinical and research workflows, investment in scalable, structured real-world evidence generation will be as important as continued advances in model architecture and prompt engineering.

# Figures

## Figure 1 - Evidence Badge Generation Workflow

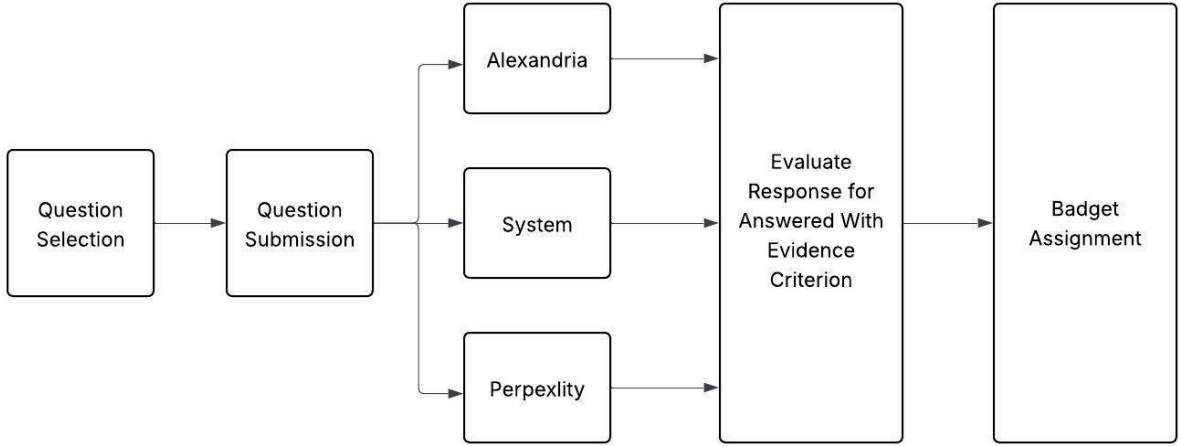

## Figure 2 - Example Context Evaluation and Badge Assignment

**Question**: In patients with heart failure, do patients with a stimulant use disorder have a difference in all cause emergency department visits, all cause 30-day readmission rates, all-cause mortality compared to patients without a stimulant use disorder?

**Atropos Response**: "Patients with heart failure and a stimulant use disorder have consistently higher all-cause emergency department visits compared to those without the disorder. For example, one study showed a significant increase in ED visits (OR 3.24, p<0.001) (Source ID: 71e2448), and similar findings were observed in additional analyses (Source ID: 70765849; f89d64b9). In contrast, findings for all-cause 30-day readmission rates are mixed. One study reported a significant increase in the stimulant use disorder group (OR 1.13, p=0.034) (Source ID: 71e24480) and another noted a significant association (Source ID: c93ff943), while other studies found no significant difference (p=0.138 and p=0.112, respectively; Source ID: 70765849, f89d64b9). All-cause mortality rates were not significantly different between groups, with one study showing a near-significant increase (HR 1.22, p=0.051) (Source ID: 71e2448) and others reporting no significant differences (p=0.82 and p=0.171; Source ID: 7076584, f89d64b)."

- Context Answers Question Directly: True
- Context Addresses Question Directly: True
- Answer Grounded in Context: True

Assessment: "The response directly addresses the comparison in outcomes using details provided in the context. It includes specific statistical results and p-values with proper citations for ED visits, 30-day readmission rates, and all-cause mortality across multiple studies."

Badge: Green

**Preplexity Response**- "In patients with heart failure, those with stimulant use disorders (e.g., methamphetamine or cocaine) exhibit higher rates of all-cause ED visits and 30-day readmissions compared to non-users, while all-cause mortality shows mixed evidence. Methamphetamine-associated heart failure (MethHF) patients had 64% 5-year HF readmission rates vs. 45% in non-MethHF [9], and cocaine users showed 49% higher 90-day all-cause readmissions [8]. Stimulant users also had more ED visits (33% vs. 22% at 30 days) [18]. Mortality trends varied: MethHF showed no significant difference in one study [10] but a 2.7x higher risk in another [12], while cocaine users had 21% increased mortality risk [8]. Contributing factors include comorbid psychiatric conditions and delayed care engagement [9][15][17]."

- Context Answers Question Directly: True
- Context Addresses Question Directly: True
- Answer Grounded in Context: False

Assessment: The provided context does include relevant data on differences in readmission rates, ED visits, and mortality among heart failure patients with stimulant use disorder, so it directly addresses the question and is related to it. However, the AI answer introduces specific percentages for ED visits (33% vs. 22% at 30 days) and mentions factors like comorbid psychiatric conditions and delayed care that are not supported by the provided context. Additionally, it cites a reference ([18]) that isn't available in the supplied material. Thus, while some parts of the answer reflect the context accurately, overall it is not well grounded solely in the provided information.

Badge: Yellow

Figure 3 - Current Evidence and LLM Capability Synergistic performance

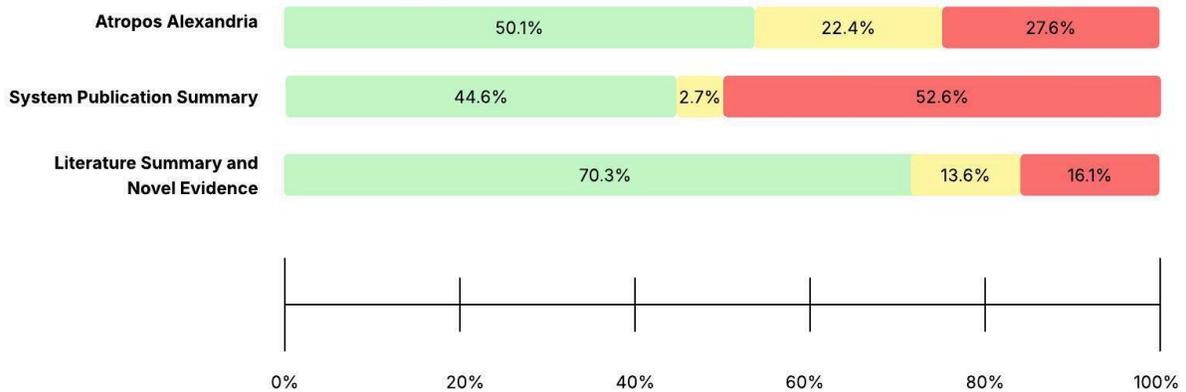

## Tables

Table 1- Answered with evidence badge summary across evidence sources.

### Table 1a - Atropos Health Summary Table

| Answered with Evidence | Total Questions | Percentage Questions |
|---|---|---|
| Green | 1,473 | 50.07 |
| Yellow | 658 | 22.37 |
| Red | 811 | 27.57 |
| Total | 2,942 | 100.0 |

### Table 1b - System Publication Summary Table

| Answered with Evidence | Total Questions | Percentage Questions |
|---|---|---|
| Green | 1,312 | 44.60 |
| Yellow | 82 | 2.79 |
| Red | 1,548 | 52.62 |
| Total | 2,942 | 100.0 |

Table 1c - Perplexity Publication Summary Table

| Answered with Evidence | Total Questions | Percentage Questions |
|---|---|---|
| Green | 371 | 21.33 |
| Yellow | 479 | 27.54 |
| Red | 889 | 51.12 |
| Total | 1,739 | 100.00 |

Table 2 - Agreement matrix between different evidence sources

Table 2a- Perplexity vs Atropos Health

| | | Perplexity Summary | | | |
|---|---|---|---|---|---|
| | | Green | Yellow | Red | Row Total |
| **Atropos Summary** | Green | 197 (11.3%) | 172 (9.9%) | 411 (23.6%) | 780 (44.9%) |
| | Yellow | 67 (3.9%) | 132 (7.6%) | 233 (13.4%) | 432 (24.8%) |
| | Red | 107 (6.2%) | 175 (10.1%) | 245 (14.1%) | 527 (30.3%) |
| | Col Total | 371 (21.3%) | 479 (27.5%) | 889 (51.1%) | 1739 (100.0%) |

Table 2b - System vs Atropos Health

| | Publication Summary | | | |
|---|---|---|---|---|
| | Green | Yellow | Red | Row Total |

| | | | | | |
|---|---|---|---|---|---|
| **Atropos Summary** | Green | 716 (24%) | 35 (1%) | 722 (25%) | 1,473 (50%) |
| | Yellow | 277 (9%) | 28 (1%) | 353 (12%) | 658 (22%) |
| | Red | 319 (11%) | 19 (1%) | 473 (16%) | 811 (28%) |
| | Col Total | 1,312 (45%) | 82 (3%) | 1,548 (53%) | 2,942 (100%) |

### Table 2c - Perplexity vs System

| | | Perplexity Summary | | | |
|---|---|---|---|---|---|
| | | Green | Yellow | Red | Row Total |
| **System Summary** | Green | 193 (11.1%) | 127 (7.3%) | 454 (26.1%) | 774 (44.5%) |
| | Yellow | 11 (0.6%) | 3 (0.2%) | 29 (1.7%) | 43 (2.5%) |
| | Red | 167 (9.6%) | 349 (20.1%) | 406 (23.3%) | 922 (53.0%) |
| | Col Total | 371 (21.3%) | 479 (27.5%) | 889 (51.1%) | 1739 (100.0%) |

### Table 2d - Agreement Across all Three Evidence Sources

| Atropos Health Badge | System Badge | Perplexity Badge | Total Cases | Percentage of Total |
|---|---|---|---|---|
| Green | Green | Green | 110 | 6.33 |
| Green | Green | Red | 218 | 12.54 |
| Green | Green | Yellow | 53 | 3.05 |
| Green | Red | Green | 81 | 4.66 |
| Green | Red | Red | 180 | 10.35 |
| Green | Red | Yellow | 119 | 6.84 |
| Green | Yellow | Green | 6 | 0.35 |
| Green | Yellow | Red | 13 | 0.75 |
| Red | Green | Green | 48 | 2.76 |
| Red | Green | Red | 115 | 6.61 |
| Red | Green | Yellow | 41 | 2.36 |
| Red | Red | Green | 56 | 3.22 |
| Red | Red | Red | 125 | 7.19 |
| Red | Red | Yellow | 132 | 7.59 |
| Red | Yellow | Green | 3 | 0.17 |
| Red | Yellow | Red | 5 | 0.29 |
| Red | Yellow | Yellow | 2 | 0.12 |
| Yellow | Green | Green | 35 | 2.01 |
| Yellow | Green | Red | 121 | 6.96 |
| Yellow | Green | Yellow | 33 | 1.9 |
| Yellow | Red | Green | 30 | 1.73 |
| Yellow | Red | Red | 101 | 5.81 |
| Yellow | Red | Yellow | 98 | 5.64 |
| Yellow | Yellow | Green | 2 | 0.12 |
| Yellow | Yellow | Red | 11 | 0.63 |
| Yellow | Yellow | Yellow | 1 | 0.06 |

# Supplement

Figure S1:

```python
prompt = f"""
# Task
You are an expert in medical research. Your job is to evaluate an AI
assistant's answer based on the provided context and question.

# Evaluation Criteria (respond True or False for each)
1. The context directly answers the question with relevant information.
2. The context is related to the question, even if it doesn't answer the
question.
3. The AI's answer is well-grounded in the provided context (no external
information or hallucinations).

# Original question
{question}

# Context provided:
{context}

# AI's answer:
{answer}

# Format
Provide your response as a structured output with 3 booleans
'quality_assessment': {{
    'context_answers_question_directly': False,
    'context_addresses_question': True,
    'answer_grounded_in_context': True,
    'assessment': 'The response accurately reflects the information available
in the context, noting the lack of direct comparison between CGM and
traditional monitoring methods regarding glycemic control and hypoglycemia
risk.'
}}
"""
```

The prompt returns a JSON-like response structured as:

```json
{
  "quality_assessment": {
    "context_answers_question_directly": true/false,
    "context_addresses_question": true/false,
    "answer_grounded_in_context": true/false,
    "assessment": "...free-text explanation..."
  }
}
```